\documentclass[10pt,twocolumn,letterpaper]{article}

\usepackage{iccv}
\usepackage{times}
\usepackage{epsfig}
\usepackage{graphicx}
\usepackage{amsmath}
\usepackage{amssymb}
\usepackage{booktabs}
\usepackage{wasysym}

\usepackage[pagebackref=true,breaklinks=true,letterpaper=true,colorlinks,bookmarks=false]{hyperref}

\def\tdmn{TMN}

\iccvfinalcopy 

\ificcvfinal\fi
\begin{document}

\title{Task-Driven Modular Networks for Zero-Shot Compositional Learning}
\author{Senthil Purushwalkam\thanks{Work done during an internship at Facebook AI Research. Proposed dataset splits and code available here: \url{http://www.cs.cmu.edu/\~spurushw/projects/compositional.html}}\\
Carnegie Mellon University\\
{\tt\small spurushw@andrew.cmu.edu}
\and 
Maximilian Nickel\\
Facebook AI Research\\
{\tt\small maxn@fb.com}
\and 
Abhinav Gupta\\
Facebook AI Research\\
{\tt\small gabhinav@fb.com}
\and 
Marc'Aurelio Ranzato\\
Facebook AI Research\\
{\tt\small ranzato@fb.com}
}

\maketitle

\begin{abstract}
\vspace{-8pt}
One of the hallmarks of human intelligence is the ability to compose learned
knowledge into novel concepts which can be recognized without a single training example. In contrast, current state-of-the-art methods require hundreds of training examples for each possible category to build reliable and accurate classifiers. To alleviate this striking difference in efficiency, we propose a task-driven modular
architecture for compositional reasoning and sample efficient learning. Our architecture consists of a set of neural network modules, which are small fully connected layers operating in semantic concept space. These modules are configured through a gating function conditioned on the task to produce features representing the compatibility between the input image and the concept under consideration. This enables us to express tasks as a combination of sub-tasks and to generalize to unseen categories by reweighting a set of small modules.
Furthermore, the network can be trained efficiently as it is fully differentiable and its modules operate on small sub-spaces. We focus our study on the problem of compositional zero-shot classification of object-attribute categories. We show in our experiments that current evaluation metrics are flawed as they only consider unseen object-attribute pairs. When extending the evaluation to the generalized setting which accounts also for pairs seen during training, we discover that na\"ive baseline methods perform similarly or better than current approaches. However, our
modular network is able to outperform all existing approaches on two widely-used benchmark
datasets.
\end{abstract}

\section{Introduction}
\label{sec:intro}

\begin{figure}[t!]
	\centering
	\includegraphics[width=\columnwidth]{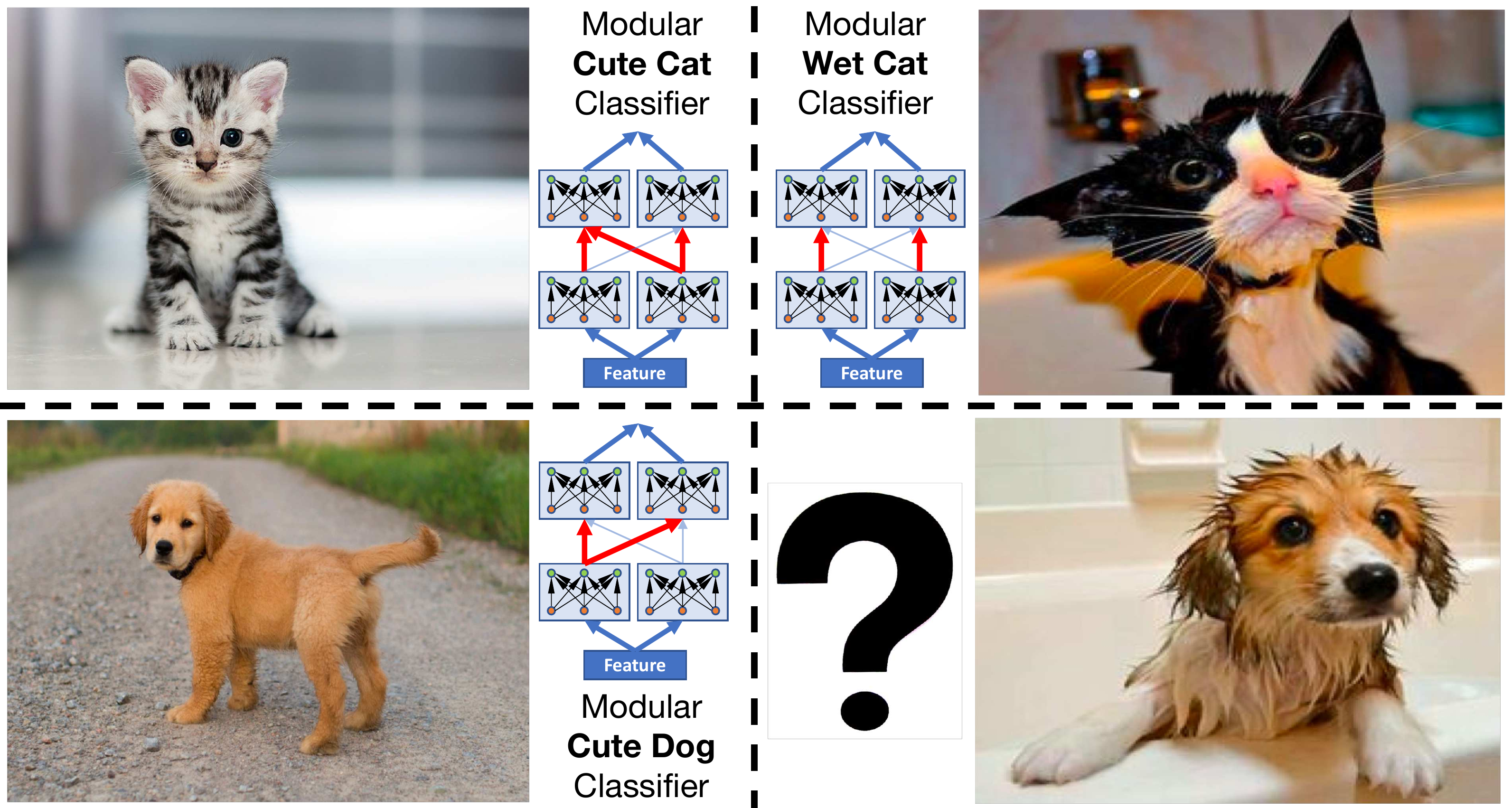}
	\caption{\small We investigate how
	to build a classifier on-the-fly, for a new concept (``wet dog") given knowledge of related concepts (``cute dog", ``cute cat", and ``wet cat"). Our approach consists of a modular network operating in a semantic feature space. By rewiring its primitive modules, the network can recognize new structured concepts.}
	\label{fig:teaser}
\end{figure}

How can machines reliably recognize the vast number of possible visual concepts?
Even simple concepts like ``envelope'' could, for instance, be divided into a
seemingly infinite number of sub-categories, e.g., by size (large, small), color (white, yellow), type (plain, windowed), or condition (new, wrinkled, stamped).
Moreover, it has frequently been observed that visual
concepts follow a long-tailed distribution \cite{salakhutdinov2011learning,wang2017learning,van2017devil}.
Hence, most classes are rare, and yet humans are able to recognize them without
having observed even a single instance. Although a surprising event, most
humans wouldn't have trouble to recognize a ``tiny striped purple elephant
sitting on a tree branch''. For machines, however, this would constitute a
daunting challenge. It would be impractical, if not impossible, to gather
sufficient training examples for the long tail of all possible categories, even
more so as current learning algorithms are data-hungry and rely on large amounts
of labeled examples. How can we build algorithms to cope with this challenge?

One possibility is to exploit the {\em compositional} nature of the prediction task. While a machine may not have observed any images of ``wrinkled envelope'', it may have observed many more images of ``white envelope'', as well as ``white paper'' and ``wrinkled paper''. If the machine is capable of compositional reasoning, it may be able to {\em transfer} the concept of being ``wrinkled'' from ``paper'' to ``envelope'', and generalize without requiring additional examples of actual ``wrinkled envelope''. 

One key challenge in compositional reasoning is {\em contextuality}. The meaning of an attribute, and even the meaning of an object, may be dependent on each other. 
For instance, how ``wrinkled'' modifies the appearance of ``envelope'' is very different from how it changes the appearance of ``dog''. In fact, contextuality goes beyond semantic categories. The way ``wrinkled'' modifies two images of ``dog'' strongly depends on the actual input dog image. In other words, the model should capture intricate interactions between {\em the image, the object and the attribute} in order to perform correct inference. While most recent approaches~\cite{misra17, Nagarajan18} 
capture the contextual relationship between object and attribute, they still rely on the original feature space being rich enough, as inference entails matching
image features to an embedding vector of an object-attribute pair.

In this paper, we focus on the task of compositional learning, where the model has to predict the object present in the input image (e.g., ``envelope''), as well as its corresponding attribute (e.g., ``wrinkled''). We believe there are two key ingredients required: (a) learning high-level sub-tasks which may be useful to transfer concepts, and 
(b) capturing rich interactions between the image, the object and the attribute. In order to capture both these properties, we propose {\bf Task-driven Modular Networks} (\tdmn).

First, we tackle the problem of transfer and re-usability by employing modular networks in the high-level semantic space of CNNs~\cite{lecun98, he16}. 
The intuition is that by modularizing in the concept space, modules can now represent common high-level sub-tasks over which ``reasoning'' can take place: 
in order to recognize a new object-attribute pair, the network simply re-organizes its computation on-the-fly by appropriately reweighing  modules for the new task. 
Apart from re-usability and transfer, modularity has additional benefits: (a) sample efficiency: transfer reduces to figuring out how to gate modules, 
as opposed to how to learn their parameters; (b) computational efficiency: since modules operate in smaller dimensional sub-spaces, 
predictions can be performed using less compute; and (c) interpretability: as modules specialize and similar computational paths are used for visually similar pairs, 
users can inspect how the network operates to understand which object-attribute pairs are deemed similar, which attributes drastically change appearance, etc. (\textsection\ref{subsec:quantitative}). 

Second, the model extracts features useful to assess the {\em joint-compatibility} between the input image and the object-attribute pair. While prior work~\cite{misra17,Nagarajan18} mapped images in the embedding space of objects and attributes by extracting features only based on images, our model instead extracts features that depend on all the members of the input triplet. The input object-attribute pair is used to rewire the modular network 
to ultimately produce features {\em invariant} to the input pair. While in prior work the object 
and attribute can be extracted from the output features, in our model features are exclusively optimized to discriminate the validity of the input triplet.

Our experiments in \textsection\ref{sec:exp_quant} demonstrate that our approach outperforms all previous approaches under the ``generalized'' evaluation protocol 
on two widely used evaluation benchmarks. The use of the generalized evaluation protocol, which tests performance on both unseen {\em and seen} pairs, gives a more precise understanding of the generalization ability of a model~\cite{chao16}. In fact, we found that under this evaluation protocol 
baseline approaches often outperform the current state of the art. Furthermore, our qualitative analysis shows that our fully differentiable modular network learns 
to cluster together similar concepts and has intuitive interpretation.
We will publicly release the code and models associated with this paper.

\section{Related Work} \label{sec:related}
Compositional zero-shot learning (CZSL) is a special case of zero-shot learning (ZSL)~\cite{palatucci09, lampert14}. In ZSL the learner
observes input images and corresponding class descriptors. Classes seen at test time never overlap with classes seen at training time,
and the learner has to perform a prediction of an unseen class by leveraging its class descriptor without any training image (zero-shot). 
In their seminal work, Chao et al.~\cite{chao16} showed that ZSL's evaluation methodology is severely limited because it only accounts for performance on unseen classes,
and they propose i) to test on both seen and unseen classes (so called ``generaralized'' setting) and ii) to calibrate models to strike the best trade-off between achieving a good performance
on the seen set and on the unseen set. In our work, we adopt the same methodology and calibration technique, although alternative calibration techniques have also been explored in literature~\cite{Cacheux19,Liu18}.
The difference between our generalized CZSL (GCZSL) setting and generalized ZSL is that we predict not only an object id, but also its corresponding attribute.
The prediction of such pair makes the task compositional as given $N$ objects and $M$ attributes, there are potentially $N*M$ possible pairs the learner could predict.

Most prior approaches to CZSL are based on the idea of embedding the object-attribute pair in image feature space~\cite{misra17,Nagarajan18}. In our work instead, we propose to learn the joint compatibility~\cite{lecun_tutorial06} between the input image and the pair by learning a representation that depends on the input triplet, 
as opposed to just the image. This is potentially more expressive as it can capture intricate dependencies between image and object-attribute pair.

A major novelty compared to past work is also the use of modular networks. Modular networks can be interpreted as a generalization of hierarchical mixture of 
experts~\cite{moe,hmoe,eigen14}, where each module holds a distribution over all the modules at the layer below and where the gatings do not depend on the input image but on a task descriptor. 
These networks have been used in the past to speed up computation at test time~\cite{Ahmed18} 
and to improve generalization for multi-task learning~\cite{Meyerson18, rosenbaum18}, reinforcement learning~\cite{fernando17}, continual learning~\cite{Valkov18}, 
visual question answering~\cite{andreas16, film18}, etc. but never for CZSL.

The closest approach to ours is the concurrent work by Wang et al.~\cite{wang18}, where the authors factorize convolutional layers and perform a component-wise gating which depends on
the input object-attribute pair, therefore also using a task driven architecture. 
This is akin to having as many modules as feature dimensions, which is a form of degenerate modularity since individual feature dimensions are unlikely to model high-level sub-tasks. 

Finally, our gating network which modulates the computational blocks in the recognition network, can also be interpreted as a particular instance of 
meta-learning~\cite{metalearning, matchingnet}, whereby the gating network predicts on-the-fly a subset of task-specific parameters (the gates) in the recognition network.

\section{Approach} \label{sec:approach}
\begin{figure*}[t!]
	\centering
	\includegraphics[width=0.7\textwidth]{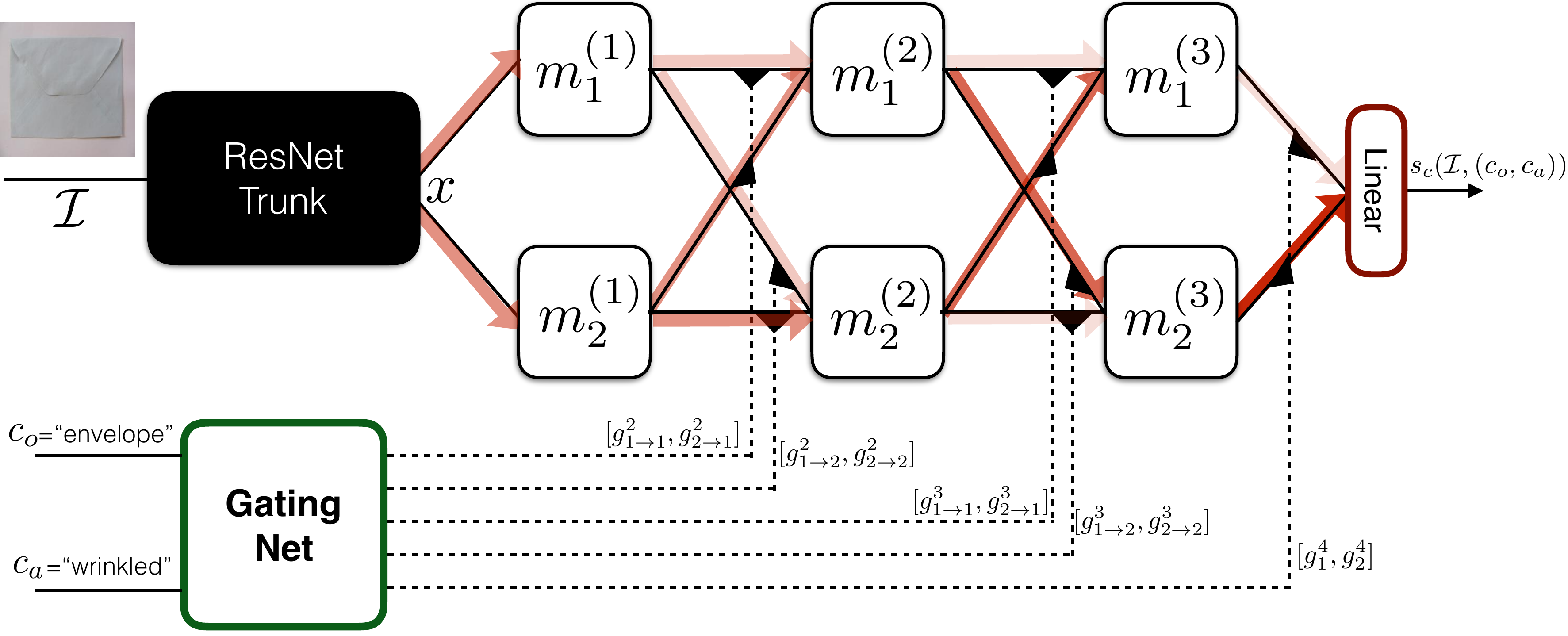}
	\caption{\small Toy illustration of the task-driven modular network (\tdmn). A pre-trained ResNet trunk extracts high level semantic representations of an input image.
These features are then fed to a modular network (in this case, three layers with two modules each) whose blocks are gated (black triangle amplifiers) by a gating network. 
The gating network takes as input an object and an attribute id. Task driven features are then projected into a single scalar value representing 
the joint compatibility of the triplet (image, object and attrtibute). The overlaid red arrows show the strength of the gatings on each edge.}
	\label{fig:model}
\end{figure*}

Consider the visual classification setting where each image $\mathcal{I}$ is associated with a visual concept $c$. 
The manifestation of the concepts $c$ is highly structured in the visual world. In this work, we consider the setting where images are the composition
of an object (e.g., ``envelope'') denoted by $c_o$, and an attribute (e.g., ``wrinkled'') denoted by $c_a$; therefore, $c = (c_o, c_a)$. 
In a fully-supervised setting, classifiers are trained for each concept $c$ using a set of human-labelled images and then tested on novel images belonging 
to the same set of concepts. Instead, in this work we are interested
in leveraging the compositional nature of the labels to extrapolate classifiers to novel concepts at test time, even without access to any training examples 
on these new classes (zero-shot learning).

More formally, we assume access to a training set $\mathcal{D}_\text{train} = \{(\mathcal{I}^{(k)}, c^{(k)}) ~ | ~ k=1,2,...,N_\text{train}\}$ 
consisting of image $\mathcal{I}$ labelled with a concept $c \in \mathcal{C}_\text{train}$, with 
$\mathcal{C}_\text{train} \subset \mathcal{C}_o \times \mathcal{C}_a = \{ (c_o, c_a)  ~ | ~ c_o \in \mathcal{C}_o, c_a \in \mathcal{C}_a\}$ 
where $\mathcal{C}_o$ is the set of objects and $\mathcal{C}_a$ is the set of attributes.

In order to evaluate the ability of our models to perform zero-shot learning, we use a similar validation ($\mathcal{D}_\text{val}$) and test 
($\mathcal{D}_\text{test}$) sets consisting of images labelled with concepts from $\mathcal{C}_\text{val}$ and $\mathcal{C}_\text{test}$, respectively. 
In contrast to a fully-supervised setting,  validation and test concepts do not fully overlap with training concepts, 
i.e. $\mathcal{C}_\text{val} \backslash \mathcal{C}_\text{train} \neq \emptyset$, $\mathcal{C}_\text{test} \backslash \mathcal{C}_\text{train} \neq \emptyset$ and $\mathcal{C}_\text{cal} \cap \mathcal{C}_\text{train} \neq \emptyset$, $\mathcal{C}_\text{test} \cap \mathcal{C}_\text{train} \neq \emptyset$. 
Therefore, models trained to classify training concepts must also generalize to ``unseen'' concepts to successfully classify images in the validation and test sets.
We call this learning setting, {\em Generalized} Zero-Shot Compositional learning, as both seen and unseen concepts appear in the validation and test sets. Note that this setting is unlike
standard practice in prior literature where a common validation set is absent and only unseen pairs are considered in the test set~\cite{misra17,Nagarajan18,wang18}.

In order to address this compositional zero-shot learning task, we propose a Task-driven Modular Network (\tdmn) which we describe next.

\subsection{Task-Driven Modular Networks (\tdmn)} \label{subsec:gated_mod_nets}
The basis of our architecture design is a {\em scoring model}~\cite{lecun_tutorial06} of the joint compatibility between image, object and attribute. This is motivated by the fact that each member of the triplet exhibits intricate dependencies with the others, i.e.  
how an attribute modifies appearance depends on the object category as well as the specific input image.
Therefore, we consider a function that takes as input the whole triplet and extracts representations of it in order to assign a compatibility score.
The goal of training is to make the model assign high score to correct triplets (using the provided labeled data), and low score to incorrect triplets.
The second driving principle is {\em modularity}. Since the task is compositional, we add a corresponding inductive bias by using a modular network. During training 
the network learns to decompose each recognition task into sub-tasks that can then be combined in novel ways at test time, consequently yielding generalizeable classifiers.

The overall model is outlined in Fig.~\ref{fig:model}. It consists of two components: a gating model $\mathcal{G}$ and a feature extraction model $\mathcal{F}$. The latter $\mathcal{F}$ consists of a set of neural network modules, which are small, fully-connected layers but could be any other parametric differentiable function as well. These modules are used on top of a standard ResNet pre-trained trunk. Intuitively, the ResNet trunk is used to map the input image $\mathcal{I} $ to a semantic concept space where higher level ``reasoning'' can be performed. We denote the mapped $\mathcal{I}$ in such semantic space with $x$. 
The input to each module is a weighted-sum of the outputs of all the modules at the layer below, with weights determined by the gating model $\mathcal{G}$, which 
effectively controls how modules are composed. 

Let $L$ be the number of layers in the modular part of $\mathcal{F}$, $M^{(i)}$ be the number of modules in the $i$-th layer, $m^{(i)}_j$
be $j$-th module in layer $i$ and $x^{(i)}_j$ be the input to each module\footnote{We set $o^{(0)}_1 = x $, $M^{(0)} = 1$, and $M^{(L)} = 1$.}, then we have:
\begin{gather}
\label{eq:module_input}
  x^{(i)}_j = \sum_{k=1}^{M^{(i-1)}} g^{(i)}_{k\rightarrow j} * o^{(i-1)}_k,
\end{gather}
where $*$ is the scalar-vector product, the output of the $k$-th module in layer $(i-1)$ is $o^{(i-1)}_k = m^{(i-1)}_k\big[x^{(i-1)}_k\big]$ 
and the weight on the edge between $m^{(i-1)}_k$ and $m^{(i)}_j$ is denoted by $g^{(i)}_{k\rightarrow j} \in \mathbb{R}$. 
The set of gatings $g = \{g^{(i)}_{k\rightarrow j} ~ | ~ i \in [1,L], j\in[1, M^{(i)}], k\in[1, M^{(i-1)}] \}$ jointly represent how modules are composed for scoring
a given concept.

The gating network $\mathcal{G}$ is responsible for producing the set of gatings $g$ given 
a concept $c=(c_o,c_a)$ as input. $c_o$ and $c_a$ are represented as integer ids\footnote{Our framwork can be trivially extended to the case where  $c_o$ and $c_a$ are structured, e.g., word2vec vectors~\cite{mikolov_arxiv13}. This would enable generalization not only to novel combinations of existing objects and attributes, but also
to novel objects and novel attributes.}, and are then embedded using a learned lookup table. These embeddings are then concatenated and processed by a multilayer neural network which computes the gatings as:
\begin{gather}
\label{eq:gating_out}
\mathcal{G}(c) = [q^{(1)}_{1\rightarrow 1}, q^{(1)}_{2\rightarrow 1}, .... q^{(L)}_{ M^{(L-1)}\rightarrow M^{(L)}}], \\
g^{(i)}_{k\rightarrow j} = \frac{\exp[q^{(i)}_{k\rightarrow j}]}{\sum_{k'=1}^{M^{(i-1)}} \exp[q^{(i)}_{k'\rightarrow j}]}.
\end{gather}
Therefore, all incoming gating values to a module are positive and sum to one.

The output of the feature extraction network $\mathcal{F}$ is a feature vector, $o^{(L)}_1$, which is
linearly projected into a real value scalar to yield the final score, $s_c(\mathcal{I}, (c_o, c_a))$. This represents the compatibility of the input triplet, see Fig.~\ref{fig:model}.  

\subsection{Training \& Testing}
\label{subsec:training}
Our proposed training procedure involves jointly learning the parameters of both gating and feature extraction networks 
(without fine-tuning the ResNet trunk for consistency with prior work~\cite{misra17,Nagarajan18}). 
Using the training set described above, for each sample image $\mathcal{I}$ we compute scores for all concepts $c =(c_o, c_a) \in \mathcal{C}_\text{train}$ 
and turn scores into normalized probabilities with a softmax: $p_c = \frac{\exp[s_c]}{\sum_{c'\in C_\text{train}} \exp[s_{c'}]}$. 
The standard (per-sample) cross-entropy loss is then used to update the parameters of both $\mathcal{F}$ and $\mathcal{G}$: ${\mathcal{L}(\mathcal{I}, \hat{c}) = - \log p_{\hat{c}}}$, if
$\hat{c}$ is the correct concept.

In practice, computing the scores of all concepts may be computationally too expensive if $\mathcal{C}_\text{train}$ is large. 
Therefore, we approximate the probability normalization factor by sampling a random subset of negative candidates~\cite{bengio03}. 

Finally, in order to encourage the model to generalize to unseen pairs, we regularize using a method we dubbed {\em ConceptDrop}. At each epoch, we choose a small random subset of pairs, exclude those samples and also do not consider them for negative pairs candidates. We cross-validate the size of the ConceptDrop subset for all the models. 

At test time, given an image we score all pairs present in $\mathcal{C}_\text{test} \cup \mathcal{C}_\text{train}$, and select the pair yielding the largest score. However, often the model is not calibrated for unseen concepts, since the unseen concepts were not involved in the optimization of the model. Therefore, we could add a scalar bias term to the score of any unseen concept~\cite{chao16}. Varying the bias from very large negative values to very large positive values has the overall effect of limiting classification to only seen pairs or only unseen pairs respectively. Intermediate values strike a trade-off between the two. 
\section{Experiments} \label{sec:exp}
We first discuss datasets, metrics and baselines used in this paper. We then report our experiments on two widely used benchmark datasets for CZSL, and we conclude with a qualitative analysis demonstrating how \tdmn~operates. Data and code will be made publicly available.

\paragraph{Datasets} We considered two datasets. The {\bf MIT-States} dataset~\cite{mitstates} has 245 object classes, 115 attribute classes and about 53K images.
On average, each object is associated with 9 attributes. There are diverse object categories, such as ``highway'' and ``elephant'', and there is also large variation in the attributes, e.g. ``mossy''
and ``diced'' (see Fig.~\ref{fig:mit_states_embedding} and \ref{fig:res_retrieval} for examples). The training set has about 30K images belonging to 
1262 object-attribute pairs (the {\em seen} set), the validation set has about 10K images from 300 seen and 300 unseen pairs, 
and the test set has about 13K images from 400 seen and 400 unseen pairs.

The second dataset is {\bf UT-Zappos50k}~\cite{zappos} which has 12 object classes and 16 attribute classes, with a total of about 33K images. This dataset consists of
different types of shoes, e.g. ``rubber sneaker'', ``leather sandal'', etc. and requires fine grained classification ability. 
This dataset has been split into a training set containing about 23K images from 83 pairs (the seen pairs),
a validation set with about 3K images from 15 seen and 15 unseen pairs, and a test set with about 3K images from 18 seen and 18 unseen pairs.

The splits of both datasets are different from those used in prior work~\cite{Nagarajan18,misra17}, now allowing fair cross-validation of hyperparameters and evaluation in the \textit{generalized} zero-shot learning setting. We will make the splits publicly available to facilitate easy comparison for future research.

\paragraph{Architecture and Training Details}
The common trunk of the feature extraction network is a ResNet-18~\cite{he16} pretrained on ImageNet~\cite{russakovsky2014imagenet} which is not finetuned,
similar to prior work~\cite{misra17,Nagarajan18}. Unless otherwise stated, our modular network has 24 modules in each layer. Each module operates in a 16 dimensional space, i.e. the dimensionality of $x^{(i)}_j$ and $o^{(i)}_j$ in eq.~\ref{eq:module_input} is 16. 
Finally, the gating network is a 2 layer neural network with 64 hidden units. The input lookup table is initialized with Glove word embeddings~\cite{pennington_emnlp14} as in prior 
work~\cite{Nagarajan18}. The network is optimized by stochastic gradient descent with ADAM~\cite{adam} with minibatch size equal to 256. 
All hyper-parameters are found by cross-validation on the validation set (see \textsection\ref{sec:ablation} for robustness to number of layers and number of modules). 
\paragraph{Baselines}
We compare our task-driven modular network against several baseline approaches. First, we consider the {\em RedWine} method~\cite{misra17} which represents objects and attributes
via SVM classifier weights in CNN feature space, and embeds these parameters in the feature space to produce a composite classifier for the (object, attribute) pair.
Next, we consider {\em LabelEmbed+}~\cite{Nagarajan18} which is a common compositional learning baseline. This model involves embedding the concatenated (object, attribute) Glove word vectors and the ResNet feature of an image, into a joint feature space using two separate multilayer neural networks. Finally, we consider the recent {\em AttributesAsOperators} approach~\cite{Nagarajan18}, 
which represents the attribute with a matrix and the 
object with a vector. The product of the two is then multiplied by a projection of the ResNet feature space to produce a scalar score of the input triplet. All methods use the same ResNet features as ours. Note that architectures from \cite{misra17, Nagarajan18} have more parameters compared to our model. Specifically, RedWine, LabelEmbed+ and AttributesAsOperators have approximately 11, 3.5 and 38 times more parameters (excluding the common ResNet trunk) than the proposed \tdmn. 

\paragraph{Metrics}
We follow the same evaluation protocol introduced by Chao et al.~\cite{chao16} in {\em generalized} zero-shot learning, 
as all prior work on CZSL only tested performance on unseen pairs without controlling accuracy on seen pairs. 
Most recently, Nagarajan et al.~\cite{Nagarajan18} introduced an ``open world'' setting whereby both seen and unseen pairs are considered during scoring but 
only unseen pairs are actually evaluated. As pointed out by Chao et al.~\cite{chao16}, this methodology is flawed because, depending on how the system is trained,
seen pairs can evaluate much better than unseen pairs (typically when training with cross-entropy loss that induces negative biases for unseen pairs) or much worse
(like in \cite{Nagarajan18} where unseen pairs are never used as negatives when ranking at training time, resulting in an implicit positive bias towards them).
Therefore, for a given value of the calibration bias (a single scalar added to the score of all unseen pairs, see \textsection\ref{subsec:training}), we compute the
accuracy on both seen and unseen pairs, (recall that our validation and test sets have equal number of 
both). 
As we vary the value of the calibration bias we draw a curve and then report its area (AUC)
later
to describe the overall performance of the system.

For the sake of comparison to prior work, 
we also report the ``closed-world'' accuracy~\cite{Nagarajan18, misra17}, i.e.
the accuracy of unseen pairs when considering only unseen pairs as candidates. 

\subsection{Quantitative Analysis} \label{sec:exp_quant}
\begin{table*}[t]
\caption{\small AUC (multiplied by 100) for MIT-States and UT-Zappos. Columns correspond to AUC computed using precision at k=1,2,3.}
    \centering
    \small
    \begin{tabular}{l c  c  c c  c  c c  c  c c c c}
    \toprule
    & \multicolumn{6}{c}{\bf MIT-States} &  \multicolumn{6}{c}{\bf UT-Zappos} \\
    \cmidrule(lr){2-7} \cmidrule(r){8-13}
    & \multicolumn{3}{c}{Val AUC} & \multicolumn{3}{c}{Test AUC} & \multicolumn{3}{c}{Val AUC} & \multicolumn{3}{c}{Test AUC} \\
    \cmidrule(lr){2-4}\cmidrule(lr){5-7}\cmidrule(lr){8-10}\cmidrule(lr){11-13}
    Model \hspace{.2cm}  Top $k\rightarrow$ & 1 & 2 & 3 & 1 & 2 & 3 & 1 & 2 & 3 & 1 & 2 & 3 \\
    \cmidrule(lr){1-1}\cmidrule(lr){2-4}\cmidrule(lr){5-7}\cmidrule(lr){8-10}\cmidrule(lr){11-13}
    AttrAsOp~\cite{Nagarajan18} &  2.5 & 6.2  & 10.1  & 1.6 & 4.7 & 7.6  & 21.5 & 44.2 & 61.6 & 25.9 & 51.3 & 67.6 \\
    RedWine~\cite{misra17} & 2.9 & 7.3 & 11.8 & 2.4 & 5.7 & 9.3  & 30.4 & 52.2 & 63.5 & 27.1 & 54.6 & 68.8 \\
    LabelEmbed+~\cite{Nagarajan18} & 3.0 & 7.6 & 12.2 & 2.0 & 5.6 & 9.4  & 26.4 & 49.0 & 66.1 & 25.7 & 52.1 & 67.8  \\
    \cmidrule(lr){1-1}\cmidrule(lr){2-4}\cmidrule(lr){5-7}\cmidrule(lr){8-10}\cmidrule(lr){11-13}
    \tdmn{} (ours) & {\bf 3.5} & {\bf 8.1} & {\bf 12.4} & {\bf 2.9}  & {\bf 7.1}  & {\bf 11.5} & \textbf{36.8} & \textbf{57.1} & \textbf{69.2} & \textbf{29.3} & \textbf{55.3} & \textbf{69.8} \\
    \bottomrule
\end{tabular}
\label{tab:auc}
\end{table*}

\begin{table}[t]
\caption{\small Best seen and unseen accuracies, and best harmonic mean of the two. See companion Fig.~\ref{fig:res_acc_curve} for the operating points used.}
    \centering
    \small
    \resizebox{\ifdim\width>\columnwidth
        \columnwidth
      \else
        \width
      \fi}{!}{
    \begin{tabular}{l c c c  c  c  c }
    \toprule
    & \multicolumn{3}{c}{\bf MIT-States} &  \multicolumn{3}{c}{\bf UT-Zappos} \\
    \cmidrule(lr){2-4}\cmidrule(lr){5-7}
    Model \hspace{.2cm}  & Seen (\Circle) & Unseen ($\times$) & HM ($\blacklozenge$) & Seen & Unseen & HM\\
    \cmidrule(lr){1-1}\cmidrule(lr){2-4}\cmidrule(lr){5-7}
    AttrAsOp & 14.3 & 17.4 & 9.9 & 59.8 & 54.2 & 40.8 \\
    RedWine &  20.7 & 17.9 & 11.6 & 57.3 & 62.3 & 41.0\\
    LabelEmbed+ & 15.0 & 20.1 & 10.7 & 53.0 & 61.9 & 40.6 \\
    \cmidrule(lr){1-1}\cmidrule(lr){2-4}\cmidrule(lr){5-7}
    \tdmn{} (ours)  & 20.2  & 20.1 & {\bf 13.0} & 58.7 & 60.0 & {\bf 45.0}\\
    \bottomrule
\end{tabular}
}
\label{tab:bestacc}
\end{table}

The main results of our experiments are reported in Tab.~\ref{tab:auc}. First, on both datasets  we observe that
\tdmn{}  performs consistently better than the other tested baselines. Second, the overall absolute values of AUC are
fairly low, particularly on the MIT-States dataset which has about 2000 attribute-object pairs and lots of potentially valid pairs for a given image due to the inherent
ambiguity of the task. Third, the best runner up method is RedWine~\cite{misra17}, suprisingly followed closely by the LabelEmbed+ baseline~\cite{Nagarajan18}.

\begin{figure}[t]
        \centering
        \includegraphics[width=\columnwidth]{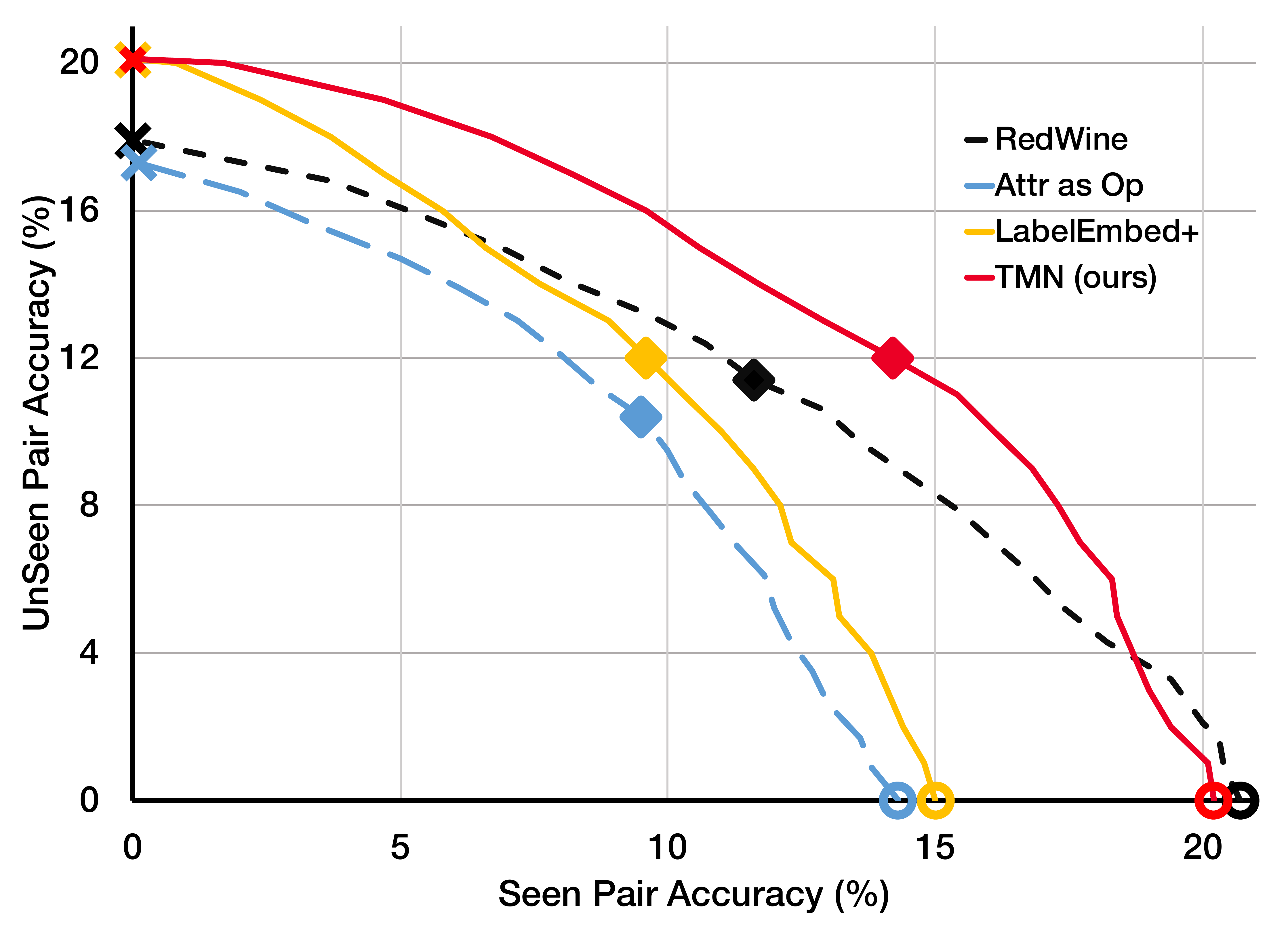}
        \caption{\small Unseen-Seen accuracy curves on MIT-States dataset. Prior work~\cite{Nagarajan18} reported unseen accuracy at different (unknown) values of seen accuracy, making comparisons inconclusive. Instead, we report AUC values~\cite{chao16}, see Tab.~\ref{tab:auc}.}
        \label{fig:res_acc_curve}
\end{figure}

The importance of using the generalized evaluation protocol becomes apparent when looking directly at the seen-unseen accuracy curve, see 
Fig.~\ref{fig:res_acc_curve}. This shows that as we increase the calibration bias we improve classification accuracy on unseen pairs but decrease the accuracy on seen pairs. Therefore, comparing methods at different operating points is inconclusive. For instance, RedWine yields the best seen pair accuracy of 20.7\% when the unseen pair 
accuracy is 0\%, compared to our approach which achieves 20.2\%, but this is hardly a useful operating point.

For the sake of comparison, we also report the best seen accuracy, the best unseen accuracy and the best harmonic mean of the two for all these methods in Tab.~\ref{tab:bestacc}.
Although our task-driven modular network may not always yield the best seen/unseen accuracy, it significantly improves the harmonic mean, indicating an overall better trade-off between the two accuracies. 

Our model not only performs better in terms of AUC but also trains efficiently. We observed that it learns from fewer updates during training.
For instance, on the MIT-States datatset, our method reaches the reported AUC of 3.5 within 4 epochs. In contrast, embedding distance based approaches such as AttributesAsOperators~\cite{Nagarajan18} and LabelEmbed+ require between 400 to 800 epochs to achieve the best AUC values using the same minibatch size. This is partly attributed to the processing of a larger number of negatives candidate pairs in each update of \tdmn (see \textsection\ref{subsec:training}). The modular structure of our network also implies that for a similar number of hidden units, the modular feature extractor has substantially fewer parameters compared to a fully-connected network. 
A fully-connected version of each layer would have $D^2$ parameters, if $D$ is the number of input and output hidden units. Instead, our modular network has $M$ blocks, each with $(\frac{D}{M})^2$ parameters. Overall, one layer of the modular network has 
$D^2 / (M * (\frac{D}{M})^2) = M$ times less parameters (which is also the amount of compute saved). See the next section for further analogies with fully connected layers.

\subsubsection{Ablation Study} \label{sec:ablation}
\begin{table}[t]
\caption{\small \textbf{Ablation study}: Top-1 valid. AUC; see \textsection\ref{sec:ablation} for details.}
    \centering
    \setlength{\tabcolsep}{1.75pt}
    \resizebox{\ifdim\width>0.9\columnwidth
        0.9\columnwidth
      \else
        \width
      \fi}{!}{
    \begin{tabular}{l c c}
    \toprule
    Model & \bf MIT-States &  \bf UT-Zappos \\
    \midrule
    \tdmn & 3.5 & 36.8\\
    a) without task driven gatings & 3.2 & 32.7 \\
    b) like a) \& no joint extraction  & 0.8 & 20.1 \\
    c) without ConceptDrop & 3.3 & 35.7\\
	\bottomrule
	\end{tabular}
    }
\label{tab:res_ablation}
\end{table}

\begin{table}[t]
\caption{\small AUC(*100) on validaton set of MIT-States varying the number of modules per layer and the number of layers.}
    \centering
    \small
    \setlength{\tabcolsep}{3pt}
    \resizebox{\ifdim\width>0.6\columnwidth
        0.6\columnwidth
      \else
        \width
      \fi}{!}{
        \begin{tabular}{c c c  c  c }
    \toprule
    & \multicolumn{4}{c}{\bf Modules} \\
    \cmidrule(l){2-5}
    \bf Layers & 12 & 18 & 24 & 30\\
    \midrule
    ~1 &1.86 & 2.14 & 2.50 & 2.51 \\
    ~3 &3.23 & 3.44 & \textbf{3.51} & 3.44\\
    ~5 &3.48 & 3.31 & 3.24 & 3.19\\
        \bottomrule
        \end{tabular}
    }
\label{tab:auc_vary_nmod}
\end{table}

In our first control experiment we assessed the importance of using a modular network by considering exactly the same architecture with two modifications. First, we learn a common set of gatings for all the concepts; henceforth, removing the task-driven modularity. And second, we feed the modular network with the concatenation of the ResNet features and the object-attribute pair embedding; henceforth, retaining the joint modeling of the triplet. To better understand this choice, consider the transformation of layer 
$i$ of the modular network in Fig.~\ref{fig:model} which can be equivalently rewritten as: 
$$
\left[
\begin{array}{c}
o^{(i)}_1 \\ 
o^{(i)}_2
\end{array}
\right] = 
\mbox{ReLU}(
\left[
\begin{array}{cc}
g^{(i)}_{1\rightarrow 1} m^{(i)}_1  & g^{(i)}_{2\rightarrow 1} m^{(i)}_1 \\
g^{(i)}_{1\rightarrow 2} m^{(i)}_2  & g^{(i)}_{2\rightarrow 2} m^{(i)}_2 
\end{array}
\right]
*
\left[
\begin{array}{c}
o^{(i-1)}_1 \\
o^{(i-1)}_2
\end{array}
\right]
)
$$
 assuming each square block $m^{(i)}_j $ is a ReLU layer. In a task driven modular network, gatings depend on the input object-attribute pair, while in this ablation study
we use gatings {\em agnostic} to the task, as these are still learned but shared across all tasks.
Each layer is a special case of a fully connected layer with a more constrained parameterization.
This is the baseline shown in row a) of Tab.~\ref{tab:res_ablation}. On both datasets performance is deteriorated showing the importance of using task driven gates. The second baseline shown in row b) of Tab.~\ref{tab:res_ablation},  is identical to the previous one but we also make the features agnostic to the task by feeding the object-attribute embedding at the {\em output} (as opposed to the input) of the modular network. This is similar to LabelEmbed+ baseline of the previous section, but replacing the fully connected layers with the same (much more constrained) architecture we use in our \tdmn{} (without task-driven gates). In this case, we can see that performance drastically drops, suggesting the importance of extracting joint representations of input image and object-attribute pair. The last row c) assesses the contribution to the performance of the ConceptDrop regularization, see  \textsection\ref{subsec:training}. Without it, AUC has a small but significant drop.

Finally, we examine the robustness to the number of layers and modules per layer in Tab.~\ref{tab:auc_vary_nmod}. Except when the modular network is very shallow, AUC is fairly
robust to the choice of these hyper-parameters.

\subsection{Qualitative Analysis} \label{subsec:quantitative}
\begin{table}[b]
\caption{\small \textbf{Edge analysis.} Example of the top 3 object-attribute pairs (rows) from MIT-States dataset that respond most strongly on 6 edges (columns) connecting blocks in the modular network.}
    \centering
    \resizebox{\ifdim\width>\columnwidth
        \columnwidth
      \else
        \width
      \fi}{!}{
    \begin{tabular}{c c c c c}
      dry river & tiny animal & cooked pasta & unripe pear & old city \\
      dry forest & small animal & raw pasta  & unripe fig & ancient city \\
      dry stream & small snake & steaming pasta & unripe apple & old town
      \end{tabular}
    }
\label{tab:edgesVSpairs}
\end{table}

\begin{table}[b]
\caption{\textbf{Module analysis.} Example of the top 3 object-attribute pairs (rows) for 6 randomly chosen modules (columns) according to the sum of outgoing edge weights in each pair's gating.}
    \centering
    \resizebox{\ifdim\width>\columnwidth
        \columnwidth
      \else
        \width
      \fi}{!}{
    \begin{tabular}{c c c c c}
      dark fire & large tree & wrinkled dress & small elephant & pureed soup \\
      dark ocean & small tree & ruffled dress  & young elephant & large pot \\
      dark cloud & mossy tree & ruffled silk & tiny elephant & thick soup
    \end{tabular}
    }
\label{tab:modulesVSpairs}
\end{table}

\begin{figure}[t]
    \centering
    \includegraphics[width=\columnwidth]{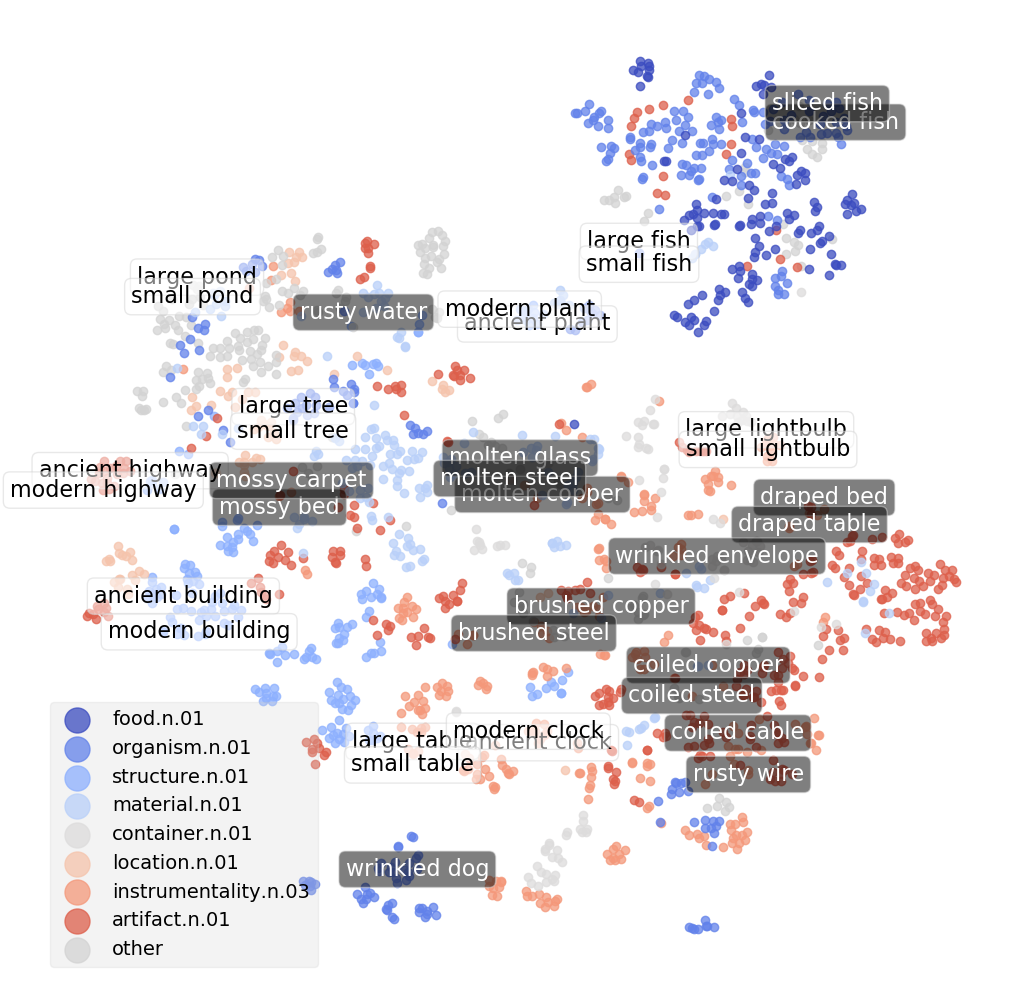}
    \caption{\small t-SNE embedding of Attribute-Object gatings on MIT-States dataset. 
 Colors indicate high-level WordNet categories of objects. 
Pairs tagged with text boxes that have white background, indicate examples where changing the attribute results in similar gatings (e.g., large/small table);
conversely, pairs in black background indicate examples where the change of attribute/object leads to very dissimilar gatings
(e.g., molten/brushed/coil steel,  rusty water/rusty wire).}
    \label{fig:mit_states_embedding}
\end{figure}

\begin{figure}[t]
        \centering
        \includegraphics[width=\columnwidth]{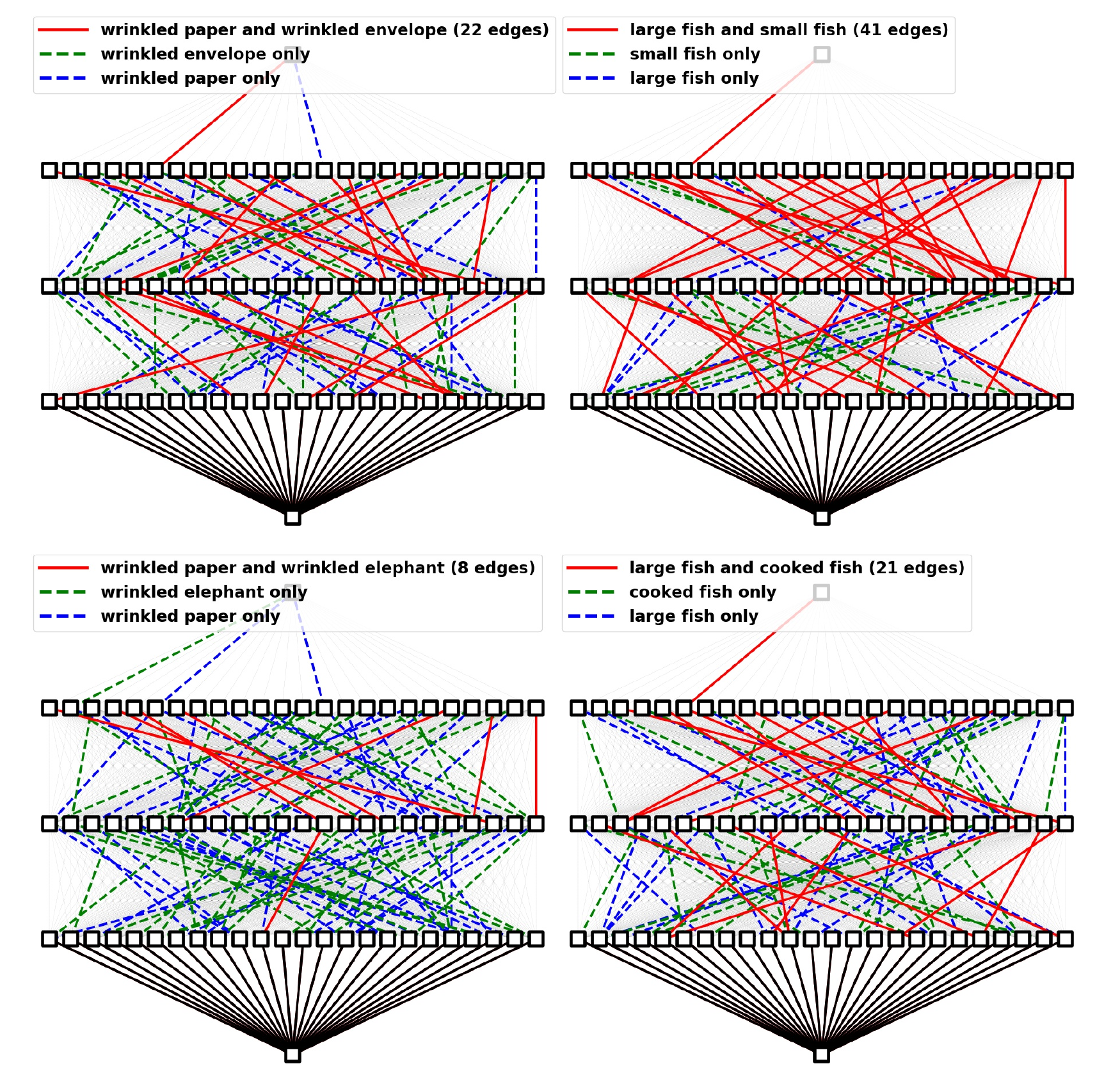}
        \caption{\small Examples of task driven topologies learned in \tdmn. Only edges whose associated weight is within 3\% of the highest weight for that edge are displayed.
        At the bottom the source features $x$, at the top the module projecting down to a single scalar scoring value.
        Each subplot compares the gatings of two object-attribute pairs.
        The red edges are the edges that are common between the two pairs. The green and the blue segments are edges acttive only in one of the two pairs. Left: Examples of
        two sets of pairs all sharing the same attribute, ``wrinkled''. Right: Examples of two sets of pairs all sharing the same object, ``fish''. Top: examples of visually
        similar pairs. Bottom: example of visually dissimilar pairs (resulting in less overlapping graphs).}
        \label{fig:res_overlapping}
\end{figure}

Task-driven modular networks are appealing not only for their performance but also because they are easy to {\em interpret}.
In this section, we explore simple ways to visualize them and inspect their inner workings. We start by visualizing the learned gatings in three ways. First, we look at which object-attribute pair has the largest gating value on a given edge of the modular network. Tab.~\ref{tab:edgesVSpairs} shows some examples indicating that visually similar pairs exhibit large gating values on the same edge of the computational graph. Similarly, we can inspect the blocks of the modular architecture. We can easily do so by associating a module to those pairs that have largest total outgoing gatings. This indicates how much a module effects the next layer for the considered pair. As shown in Tab.~\ref{tab:modulesVSpairs}, we again find that modules take ownership for explaining specific kinds of visually similar object-attribute pairs. A more holistic way to visualize the gatings is by embedding all the gating values associated with an object-attribute pair in the 2 dimensional plane using t-SNE~\cite{tsne}, as shown in Fig.~\ref{fig:mit_states_embedding}. This visualization shows that the gatings are mainly organized by visual similarity. Within this map, there are smaller clusters that correspond to the same object with various attributes. However, there are several exceptions to this, as there are instances where the attribute greatly changes the visual appearance of the object (``coiled steel" VS ``molten steel", see other examples highlighted with dark tags), for instance. Likewise, pairs sharing the same attribute may be located in distant places if the object is visually dissimilar (``rusty water" VS "rusty wire"). The last gating visualization is through the topologies induced by the gatings, as shown in Fig.~\ref{fig:res_overlapping}, where only the edges with sufficiently large gating values are shown. Overall, the degree of edge overlap 
between object-attribute pairs strongly depends on their visual similarity.

\begin{figure}[t]
    \centering
    \includegraphics[width=.7\columnwidth]{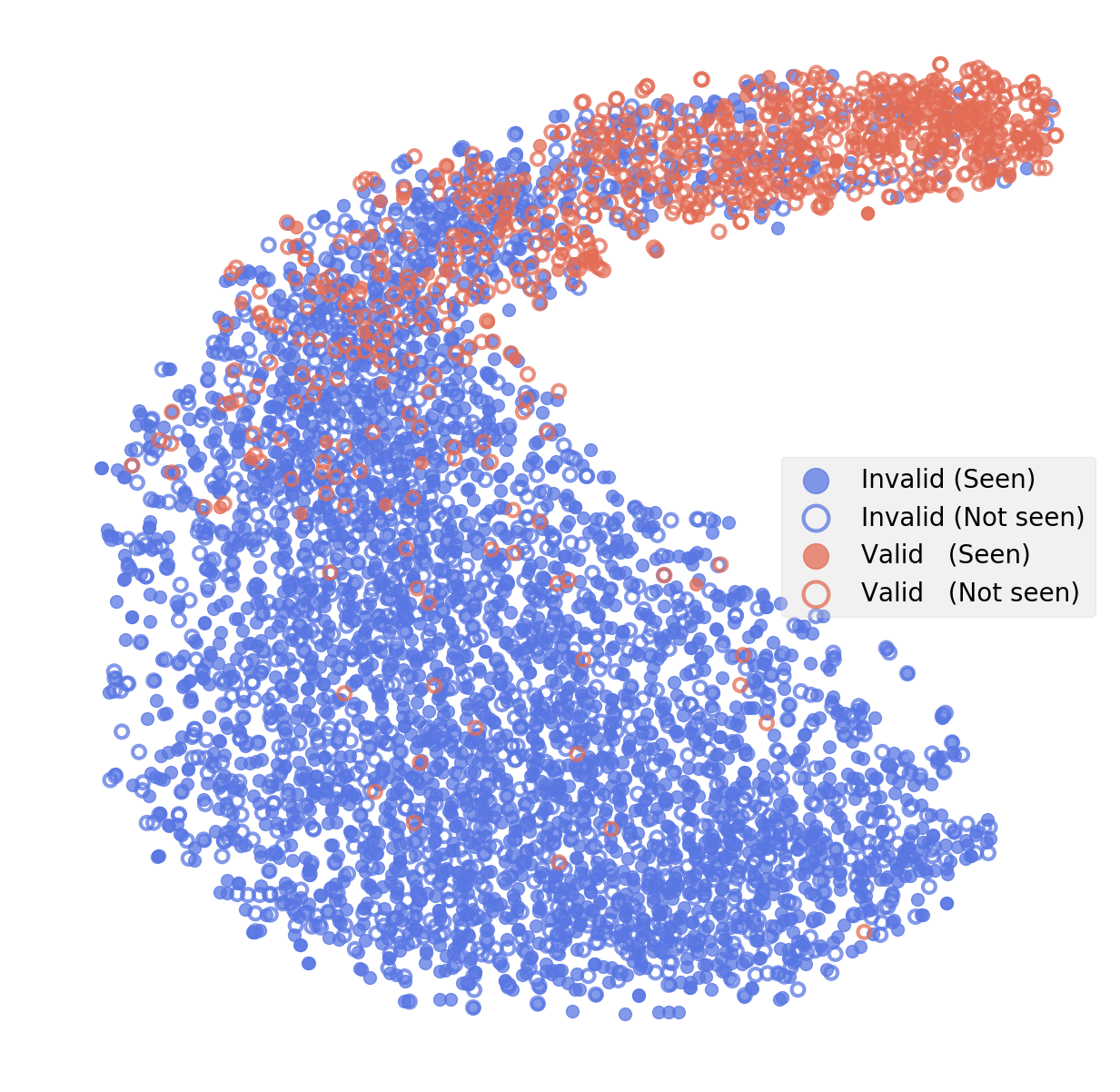}
    \caption{\small t-SNE embedding of the output features (just before the last linear projection) on MIT-States dataset. Red markers show valid (image, object, attribute) triplets (from either seen or unseen pairs), while blue markers show invalid triplets.}
    \label{fig:mit_states_feature_embedding}
\end{figure}
Besides gatings and modules, we also visualized the task-driven visual features $o^{(L)}_1$, just before the last linear projection layer, see Fig.~\ref{fig:model}. The map in Fig.~\ref{fig:mit_states_feature_embedding} shows that valid (image, object, attribute) triplets are well clustered together, while invalid triplets are nicely spread on one side of the plane. 
This is quite different than the feature organization found by methods that match concept embeddings in the image feature space~\cite{Nagarajan18,misra17}, which tend to be organized by concept. While \tdmn{} extracts largely {\em task-invariant} representations using a {\em task-driven} architecture, they produce representations that contain information about the task using a {\em task-agnostic} architecture\footnote{A linear classifier trained to predict the input object-attribute pair achieves only 5\% accuracy on \tdmn's features, and 40\% using the features of the LabelEmbed+ baseline. ResNet features obtain 41\%.}. \tdmn{} places all valid triplets on a tight cluster because the shared top linear projection layer is trained to discriminate between valid and invalid triplets (as opposed to different types of concepts). 

\begin{figure}[t]
	\centering
	\includegraphics[width=0.9\columnwidth]{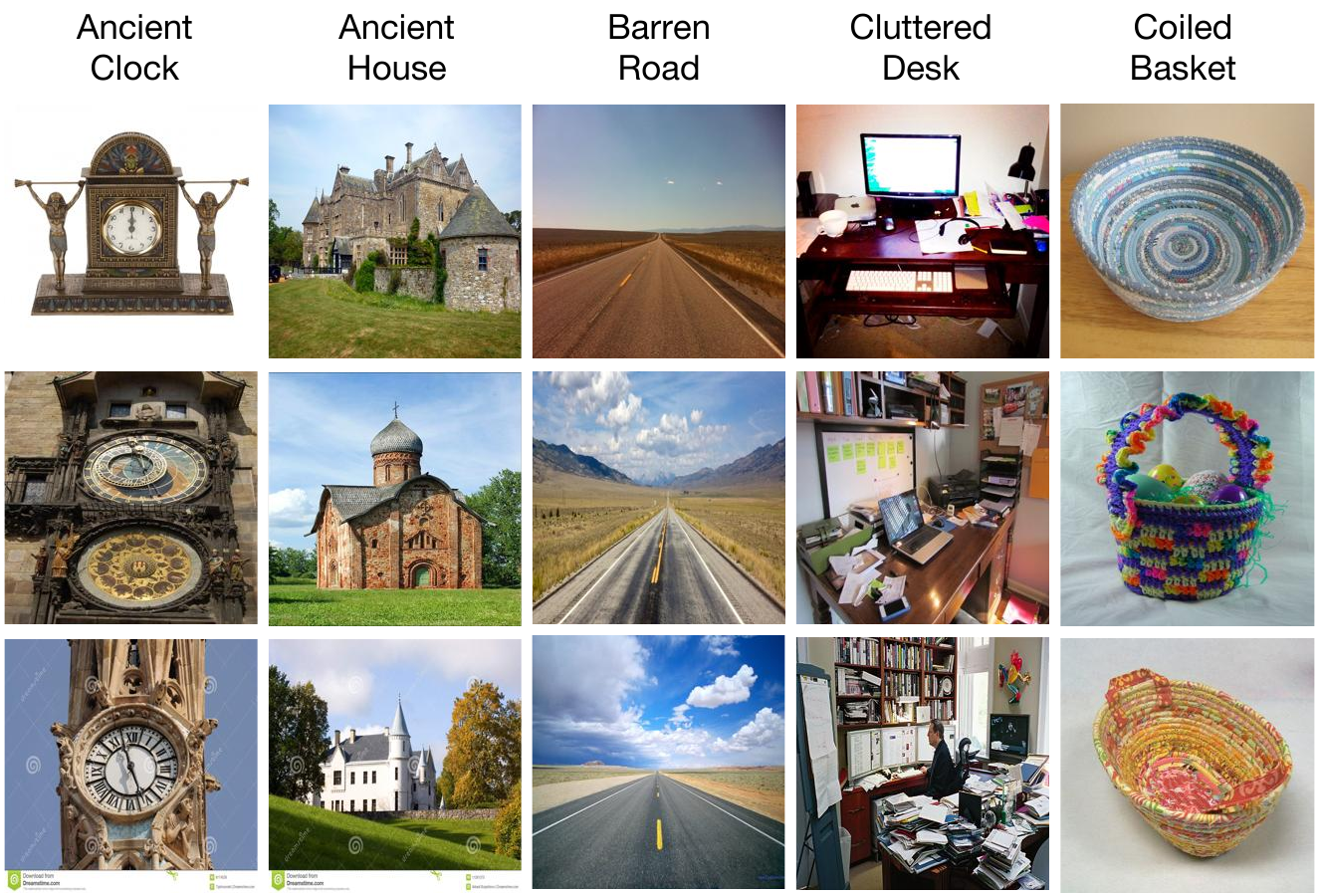}
	\caption{Example of image retrievals from the test set when querying an unseen pair (title of each column).}
	\label{fig:res_retrieval}
\end{figure}
Finally, Fig.~\ref{fig:res_retrieval} shows some retrieval results. Given a query of an unseen object-attribute pair, we rank all images from the test set (which contains both seen and unseen pairs), and return those images that rank the highest. The model is able to retrieve relevant images despite not having been exposed to these concepts during training.

\section{Conclusion} \label{sec:conclusion}
The distribution of highly structured visual concepts is very heavy tailed in nature. Improvement in sample efficiency of our current models is crucial, since labeled data will never be sufficient for concepts in the tail of the distribution.
A promising way to attack this problem is to leverage the intrinsic compositionality of the label space. In this work, we investigate this avenue of research using the Zero-Shot Compositional Learning task as a use case. 
Our first contribution is a novel architecture: \tdmn, which
outperforms all the baseline approaches we considered. There are two key ideas behind its design. First, the joint processing of input image, object and attribute to account for contextuality. And second, the use of a modular network with gatings dependent on the input object-attribute pair. Our second contribution is to advocate for the use of the generalized evaluation protocol which not only tests accuracy on unseen concepts but also seen concepts. 
Our experiments show that \tdmn{} provides better performance, while being efficient and interpretable. In future work, we will explore  other gating mechanisms and applications in other domains. 
\section{Acknowledgements}
\label{sec:acknowledge}

We would like to thank Ishan Misra, Ramakrishna Vedantam and Xiaolong Wang for the discussions and feedback that helped us in this work.

{\small
\bibliographystyle{ieee}
\bibliography{papers}
}

\appendix
\section{Hyperparameter tuning}
\label{suppsec:parameters}
The results we reported in the main paper were obtaining using the best hyper-parameters found on the validation set. We used the same cross-validation procedure for all methods, including ours. Here, we present the ranges of hyper-parameters used in the grid-search and the selected values.

\subsection{Task Driven Modular Networks}
\label{subsec:params_tdmn}

\noindent
Hyper-parameter values:
\begin{itemize}
    \item Feature extractor learning rates: 0.1, 0.01, 0.001, 0.0001 (chosen: 0.001)
    \item Gating network learning rates: 0.1, 0.01, 0.001, 0.0001 (chosen: 0.01)
    \item Number of sampled sampled negatives for Eq 3: for MIT States 200, 400, 600 (chosen: 600), for UT-Zappos we choose all negatives
    \item Batch size: 64, 128, 256, 512 (chosen: 256)
    \item Fraction of train concepts dropped in ConceptDrop: 0\%, 5\%, 10\%, 20\% (chosen: 5\%)
    \item Number of modules per layer: 12, 18, 24, 30 (chosen: 24)
    \item Output dimensions of each module: 8, 16 (chosen: 16)
    \item Number of layers: 1, 2, 3, 5 (chosen: 3 for MIT States, 2 for UT-Zappos)
\end{itemize}

\subsection{LabelEmbed+}
\label{subsec:params_labelemb}
\noindent
Hyper-parameter values:
\begin{itemize}
    \item Learning rates: 0.1, 0.01, 0.001, 0.0001 (chosen: 0.0001 for MIT States, 0.001 for UT-Zappos)
    \item Batch size: 64, 128, 256, 512 (chosen: 512)
    \item Fraction of train concepts dropped in ConceptDrop: 0\%, 5\%, 10\%, 20\% (chosen: 5\%)
\end{itemize}

\subsection{RedWine}
\label{subsec:params_redwine}
\noindent
Hyper-parameter values:
\begin{itemize}
    \item Learning rates: 0.1, 0.01, 0.001, 0.0001 (chosen: 0.01)
    \item Batch size: 64, 128, 256, 512 (chosen: 256 for MIT States, 512 for UT-Zappos)
    \item Fraction of train concepts dropped in ConceptDrop: 0\%, 5\%, 10\%, 20\% (chosen: 0\%)
\end{itemize}

\subsection{Attributes as Operators}
\label{subsec:params_attrop}
\noindent
Hyper-parameter values:
\begin{itemize}
    \item Fraction of train concepts dropped in ConceptDrop: 0\%, 5\%, 10\%, 20\% (chosen: 5\%)
\end{itemize}
Learning rate, batch size, regularization weights chosen from the original paper and executed using the implementation at: \url{https://github.com/Tushar-N/attributes-as-operators}.

\clearpage

\begin{figure*}[b!]
        \centering
        \textbf{\large 2. Additional Topology Visualizations}\par\medskip
        \includegraphics[height=0.9\textheight]{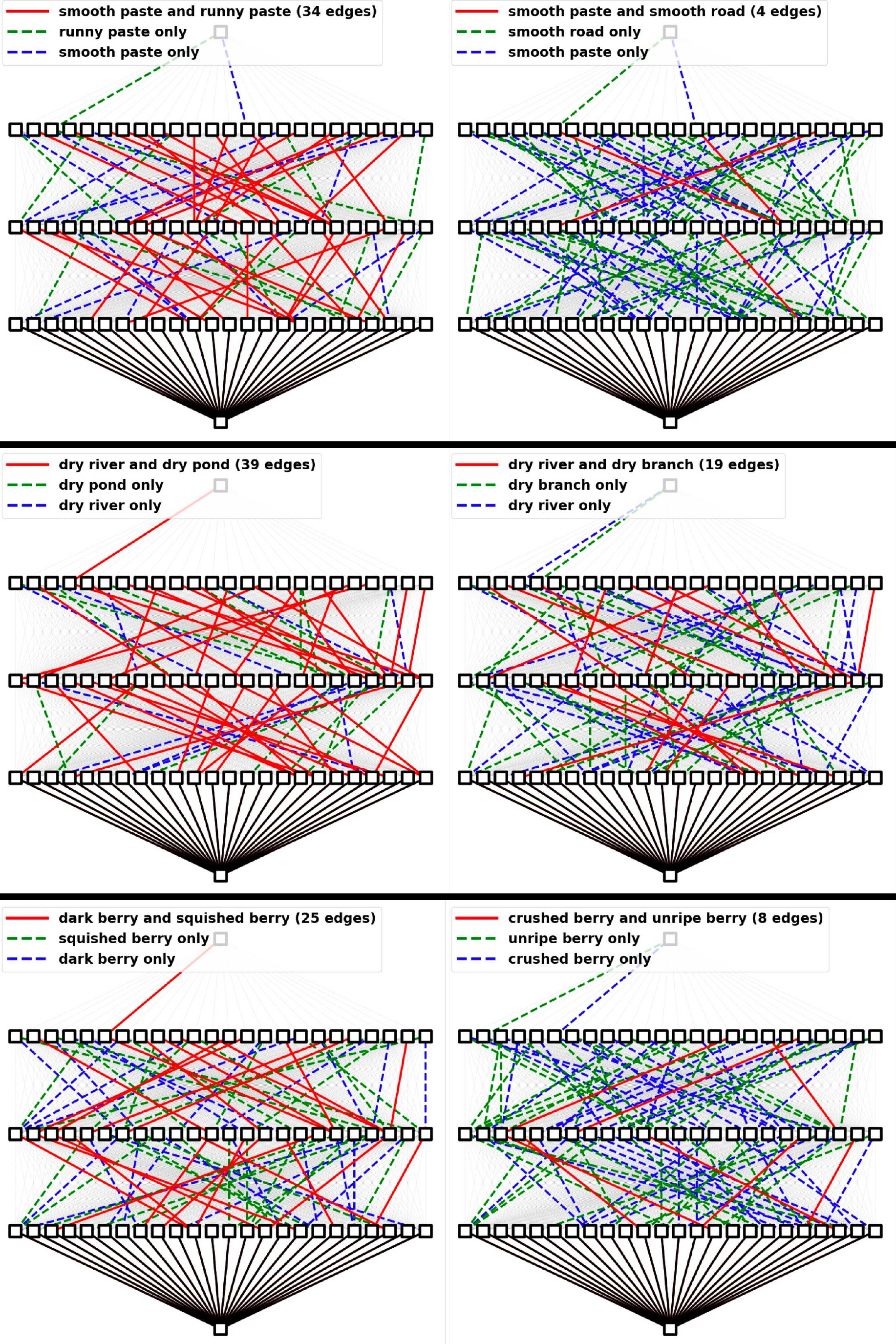}
        \caption{\small Additional Examples of task driven topologies learned in \tdmn{} (similar to Figure 5 of the main text).}
        \label{fig:res_overlapping_supp}
\end{figure*}
\clearpage

\end{document}